\documentclass{article}
\usepackage{spconf,amsmath,graphicx,cite}
\usepackage{amsthm,amssymb,mathtools,dsfont,enumitem}
\usepackage{subcaption}

\usepackage{tikz}
\usepackage[absolute]{textpos}
\newtheorem{definition}{Definition}

\newcommand{\R}[1]{\mathbb{R}^{#1}}

\newcommand{\coeff}{\boldsymbol{\alpha}}
\newcommand{\signal}{\mathbf{x}}
\newcommand{\omegab}{\boldsymbol{\omega}}

\newcommand{\yb}{\mathbf{y}}

\newcommand{\OP}{\boldsymbol{\Omega}}
\newcommand{\Ab}{\mathbf{A}}

\newcommand{\Sb}{\mathbf{S}}
\newcommand{\Ub}{\mathbf{U}}

\newcommand{\Signal}{\mathbf{X}}

\newcommand{\Acal}{\mathcal{A}}
\newcommand{\Scal}{\mathcal{S}}
\newcommand{\Xcal}{\mathcal{X}}

\newcommand{\Vcal}{\mathcal{V}}

\DeclareMathOperator{\OB}{OB}

\DeclareMathOperator{\VEC}{vec}

\newcommand{\sigdim}{n}
\newcommand{\coeffdim}{m}

\newcommand\Mark[1]{\textsuperscript{#1}}

\title{Separable Cosparse Analysis Operator Learning}
\name{Matthias Seibert\Mark{1*}, Julian W\"{o}rmann\Mark{1*}, R\'emi Gribonval\Mark{2}, Martin Kleinsteuber\Mark{1}\thanks{This work was partially supported by the German Research Foundation (DFG) under grant KL 2189/8-1, the DFG funded Cluster of Excellence CoTeSys,
and by the European Research Council, PLEASE project (ERC-StG-2011-277906).}}
\address{\Mark{1}TU M\"{u}nchen, Department of Electrical Engineering and Information Technology,
Munich, Germany.\\
\Mark{2}PANAMA Project-Team (INRIA \& CNRS).\\
\thanks{* These authors contributed equally to this work.}
\normalsize{\{m.seibert, julian.woermann, kleinsteuber\}@tum.de, gribonval@inria.fr}}
\expandafter\def\expandafter\normalsize\expandafter{%
    \normalsize
    \setlength\abovedisplayskip{5pt}
    \setlength\belowdisplayskip{5pt}
    \setlength\abovedisplayshortskip{5pt}
    \setlength\belowdisplayshortskip{5pt}
}

\begin{document}

\begin{textblock}{16}(1.915,1)
\noindent First published in the Proceedings of the 22nd European Signal Processing Conference (EUSIPCO 2014) in 2014, published by EURASIP
\end{textblock}

\maketitle

\begin{abstract}
The ability of having a sparse representation for a certain class of signals has many applications in data analysis, image processing, and other research fields. 
Among sparse representations, the cosparse analysis model has recently gained increasing interest.
Many signals exhibit a multidimensional structure, e.g.\ images or three-dimensional MRI scans.
Most data analysis and learning algorithms use vectorized signals and thereby do not account for this underlying structure. 
The drawback of not taking the inherent structure into account is a dramatic increase in computational cost.\\
We propose an algorithm for learning a cosparse Analysis Operator that adheres to the preexisting structure of the data, and thus allows for a very efficient implementation. This is achieved by enforcing a separable structure on the learned operator. Our learning algorithm is able to deal with multidimensional data of arbitrary order.
We evaluate our method on volumetric data at the example of three-dimensional MRI scans.

\end{abstract}
\section{Introduction}
\label{sec:intro}

Sparse signal models enjoy great popularity for handling various signal processing tasks such as image denoising, super-resolution, compressive sensing, and others. 
In the last decade, most of the research activity has focused on the well-known synthesis model, where the signal is synthesized via a linear combination of a few signal atoms taken from a dictionary. 
The synthesis model has a closely related counterpart called the analysis model
where signals are characterized by their sparsity in a transformed domain.
This transformation is achieved through an operator $\OP \in\R{\coeffdim \times \sigdim}$, with $\coeffdim \geq \sigdim$. When applied to a signal $\signal \in \R{\sigdim}$ it yields the coefficient vector $\coeff \in \R{\coeffdim}$ defined as
\begin{equation}
	\coeff = \OP \signal.
\end{equation}
%
In the cosparse analysis model the signal information is encoded in the zero-entries of $\coeff$.
The number of zero elements in the coefficient vector, i.e.\ $p=\coeffdim - \|\coeff\|_0$, is called the cosparsity of the signal, and we refer to $\OP$ as the cosparse analysis operator.

A prominent example for an analytic analysis operator is the finite difference operator in image processing. 
However, the advantage of the low computational complexity of such an analytically given transformation comes at the cost of a poor adaptation to specific signal classes of interest. It is now well-known that for a particular class, sparser signal representations and thus better reconstruction accuracies can be achieved if the transformation matrix is learned from a representative training set. We follow this approach in our work.
%
Various learning algorithms that aim at learning a cosparse analysis operator have been proposed, cf. \cite{yaghoobi2011analysis,nam2011cosparse,rubinstein2013analysis,hawe:tip13,Chen2014}. 
In the following, we give a short summary of the algorithms proposed in \cite{rubinstein2013analysis} and \cite{hawe:tip13}.
\cite{rubinstein2013analysis} is of importance as, to our knowledge, the only available separable cosparse analysis operator learning scheme is based on the algorithm therein, while \cite{hawe:tip13} is relevant to our cause since we extend its framework to the separable case.
The algorithm proposed in \cite{rubinstein2013analysis} is based on the famous K-SVD algorithm for dictionary learning. 
It consists of two alternating steps. First, a sparse coding step is performed, during which the current cosparse analysis operator is fixed and the best approximation of the noisy training signal with regard to the given cosparsity is computed.
Second, each row of the cosparse analysis operator is updated using only information from the training signals orthogonal to the respective row.
The general idea in \cite{hawe:tip13} is to use a geometric optimization approach in order to fulfill the requirements demanded of the cosparse analysis operator. A smooth cost function is designed that, in addition to the sparsity promoting function, includes penalty functions that ensure full rank and incoherence of the learned operator. This cost function is minimized by using a geometric conjugate gradient method.

The cosparse analysis operators produced by these learning algorithms are structured only in the sense that they have unit row norm, while no further structure is enforced. These matrices allow for a high cosparsity in the analyzed signal. However, the size of the learned operators is ultimately restricted due to limitations in memory and computational power.
Separable learning schemes tackle this problem by enforcing additional structure on the learned operators.

For the synthesis model the viability of separable approaches has been examined in \cite{rigamonti2013learning,hawe:cvpr13}. To the best of the authors' knowledge, the only work combining a separable structure with the cosparse analysis model can be found in \cite{qi:icip13}. The authors suggest a learning scheme for two-dimensional signals, which is based on the Analysis K-SVD algorithm in \cite{rubinstein2013analysis}.

We propose a separable learning scheme based on the work in \cite{hawe:tip13}.
By introducing separability we are able to deal with data of arbitrarily high order while maintaining its inherent structure. We evaluate our proposed method on volumetric Magnetic Resonance Imaging (MRI) data.\vspace{-10pt}

\paragraph*{Notation.}
Matrices are denoted as boldface capital letters such as $\Signal$. Vectors are boldface small letters, i.e.\ $\signal$. Tensors are denoted as calligraphic letters $\Xcal$, whereas the $(i_1,i_2,\ldots,i_N)$ entry of a tensor with order $N$ is denoted by $x_{i_1 i_2 \ldots i_N}$. Entries of vectors and matrices are indexed accordingly. The $i^\mathrm{th}$ row of a matrix is denoted by $\omegab_{i,:}$. Some additional notation necessary for working with tensors will be introduced in the following section.

\section{Structured Cosparse Analysis Operator Learning}
\label{sec:MAOL}

\subsection{Tensor Operations}
\begin{figure}[t]
\centering
	\begin{tikzpicture}[scale=.8]
	
	\pgfmathsetmacro{\cubex}{1}
	\pgfmathsetmacro{\cubey}{1.5}
	\pgfmathsetmacro{\cubez}{1}
	
	\pgfmathsetmacro{\cubeXx}{1.5}
	\pgfmathsetmacro{\cubeXy}{2}
	\pgfmathsetmacro{\cubeXz}{1.3}
	
	\node at (1.1,2.3,0) {$=$};
	
	\coordinate (A) at (5,3,0);
	\draw[fill=blue!20] (A) -- node[below=12pt] {$\Scal$} ++(-\cubex,0,0) -- node[left] {$I_1$} ++(0,-\cubey,0) -- node[below] {$I_2$} ++(\cubex,0,0) -- cycle;
	\draw[fill=blue!20] (A) -- ++(0,0,-\cubez) -- ++(0,-\cubey,0) -- ++(0,0,\cubez) -- cycle;
	\draw[fill=blue!20] (A) -- ++(-\cubex,0,0) -- node[above left=-2pt] {$I_3$} ++(0,0,-\cubez) -- ++(\cubex,0,0) -- cycle;
	
	\coordinate (B) at (0,3,0);
	\draw[fill=green!20] (B) -- node[below=18pt] {$\Acal$} ++(-\cubeXx,0,0) -- node[left] {$J_1$} ++(0,-\cubeXy,0) -- node[below] {$J_2$} ++(\cubeXx,0,0) -- cycle;
	\draw[fill=green!20] (B) -- ++(0,0,-\cubeXz) -- ++(0,-\cubeXy,0) -- ++(0,0,\cubeXz) -- cycle;
	\draw[fill=green!20] (B) --  ++(-\cubeXx,0,0) -- node[above left=-2pt] {$J_3$} ++(0,0,-\cubeXz) -- ++(\cubeXx,0,0) -- cycle;
	
	\draw[fill=yellow!20] (3.25,3,0) -- node[below=17pt] {$\OP^{(1)}$} node[above] {$I_1$} ++(-\cubey,0,0) -- node[left] {$J_1$} ++(0,-2,0) -- ++(\cubey,0,0) -- cycle;
	\draw[fill=yellow!20] (7.5,3,0) -- node[below=10pt] {$\OP^{(2)}$} node[above] {$I_2$} ++(-\cubex,0,0) -- node[left] {$J_2$} ++(0,-1.5,0) -- ++(\cubex,0,0) -- cycle;
	\draw[fill=yellow!20] (6.2,5,1) -- node[below right =5pt] {$\OP^{(3)}$} node[above] {$I_3$} ++(0,0,-\cubez) -- node[left=10pt] {$J_3$} ++(0,-1.3,0) -- ++(0,0,\cubez) -- cycle;
	
\end{tikzpicture}
	\caption{The $n$-mode product $\Acal = \Scal \times_1 \OP^{(1)} \times_2 \OP^{(2)} \times_3 \OP^{(3)}$ with tensors $\Acal, \Scal$ and matrices $\OP^{(i)},\, i=1,2,3$.}
\label{fig:nmode}
\end{figure}
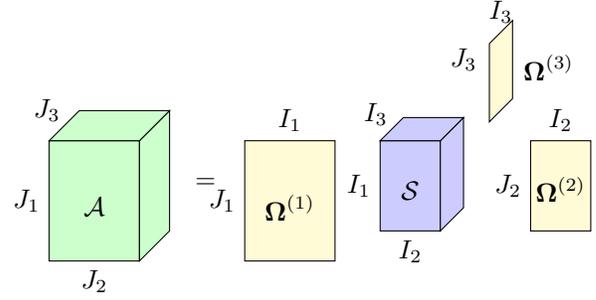

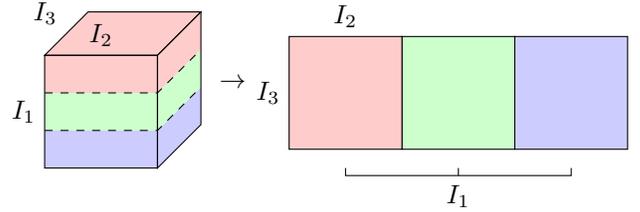
\begin{figure}[t]
\centering
\begin{tikzpicture}[scale=.5]
	\pgfmathsetmacro{\cubex}{3}
	\pgfmathsetmacro{\cubey}{3}
	\pgfmathsetmacro{\cubez}{3}
	
	%
	
	\pgfmathsetmacro{\cubeAx}{1}
	\path[fill=blue!20] (0,-2,0) -- ++(-3*\cubeAx,0,0) -- ++(0,-\cubey/3,0) -- ++(3*\cubeAx,0,0) -- cycle;
	\path[fill=blue!20] (0,-2,0) -- ++(0,0,-\cubez) -- ++(0,-\cubey/3,0) -- ++(0,0,\cubez) -- cycle;
	\path[fill=green!20] (0,-1,0) -- ++(-3*\cubeAx,0,0) -- ++(0,-\cubey/3,0) -- ++(3*\cubeAx,0,0) -- cycle;
	\path[fill=green!20] (0,-1,0) -- ++(0,0,-\cubez) -- ++(0,-\cubey/3,0) -- ++(0,0,\cubez) -- cycle;
	\path[fill=red!20] (0,0,0) -- ++(-3*\cubeAx,0,0) -- ++(0,-\cubey/3,0) -- ++(3*\cubeAx,0,0) -- cycle;
	\path[fill=red!20] (0,0,0) -- ++(0,0,-\cubez) -- ++(0,-\cubey/3,0) -- ++(0,0,\cubez) -- cycle;
	\path[fill=red!20] (0,0,0) -- ++(-3*\cubeAx,0,0) -- ++(0,0,-\cubez) -- ++(3*\cubeAx,0,0) -- cycle;

	\draw (0,0,0) -- ++(-\cubex,0,0) -- node[left] {$I_1$} ++(0,-\cubey,0) -- ++(\cubex,0,0) -- cycle;
	\draw (0,0,0) -- node[above] {$I_2$} ++(-\cubex,0,0) -- node[above left] {$I_3$} ++(0,0,-\cubez) -- ++(\cubex,0,0) -- cycle;
	\draw (0,0,0) -- ++(0,0,-\cubez) -- ++(0,-\cubey,0) -- ++(0,0,\cubez) -- cycle;
	\draw[dashed] (-3,-1,0) -- (0,-1,0) -- (0,-1,-3);
	\draw[dashed] (-3,-2,0) -- (0,-2,0) -- (0,-2,-3);
	
	\draw[fill=blue!20] (12.5,.5,0) -- ++(-\cubex,0,0) -- ++(0,-\cubey,0) -- ++(\cubex,0,0) -- cycle;
	\draw[fill=green!20] (9.5,.5,0) -- ++(-\cubex,0,0) -- ++(0,-\cubey,0) -- ++(\cubex,0,0) -- cycle;
	\draw[fill=red!20] (6.5,.5,0) -- node[above] {$I_2$} ++(-\cubex,0,0) -- node[left] {$I_3$} ++(0,-\cubey,0) -- ++(\cubex,0,0) -- cycle;
	
	\draw (5,-3,0) -- (5,-3.2,0) -- node[below] {$I_1$} (11,-3.2,0) -- (11,-3,0);
	\draw (8,-3,0) -- (8,-3.2,0);
	
	\node at (2,-.75,0) {$\rightarrow$};
\end{tikzpicture}
\caption{The 3-mode matrix unfolding of a 3-tensor.}
\label{fig:3tensor_unfolding}
\end{figure}

In order to be able to deal with multidimensional data, we have to introduce some techniques from the field of multilinear algebra. First, the so called $n$-mode product, introduced in \cite{tensor:lathauwer:2000}, provides a way to apply transformations in the form of matrices to the separate modes of the tensor.
\begin{definition}
Given the tensor $\Scal \in \R{I_1 \times I_2 \times \ldots \times I_N}$ and the matrix $\Ub \in \R{J_n \times I_n}$, the \emph{$n$-mode product} is denoted by 
\begin{equation*}
	\Scal \times_n \Ub
\end{equation*}
and results in a $I_1 \times I_2 \times \ldots \times I_{n-1} \times J_n \times I_{n+1} \times \ldots \times I_N$ tensor. The entries of this tensor are defined as
\begin{equation}
	(\Scal \times_n \Ub)_{i_1 i_2 \ldots i_{n-1} j_n i_{n+1} \ldots i_N} = \sum_{i_n = 1}^{I_n} s_{i_1 i_2 \ldots i_N} \cdot u_{j_n i_n}
\end{equation}
for all $j_n = 1,\ldots,J_n$.
\end{definition}
To offer a better understanding of this concept, a visualization for a tensor of order 3 is provided in Figure~\ref{fig:nmode}.
In addition to this product, we will also make use of the so-called \emph{$n$-mode matrix unfolding} that enables us to one-to-one map a given tensor to a matrix of appropriate size.
Since we will exclusively use unfoldings along the $N^\mathrm{th}$ mode, we will only give the definition for this specific case.
\begin{definition}
	The $N$-mode matrix unfolding of a tensor $\Scal\in \R{I_1\times \ldots \times I_N}$ is denoted by $\Sb_{(N)} \in \R{I_N \times (\prod_{j \neq N} I_j)}$. Given the tensor $\Scal \in \R{I_1 \times \ldots \times I_N}$, the unfolding matrix $\Sb_{(N)}$ contains the element $s_{i_1 i_2 \ldots i_N}$ in row $i_N$ and column
	\begin{equation*}
		(i_1 - 1)I_2 I_3 \ldots I_{N-1} + (i_2 - 1) I_3 I_4 \ldots I_{N-1} + \ldots + i_{N-1}.
	\end{equation*}
\end{definition}
For a general definition of this mapping we refer the interested reader to \cite{tensor:lathauwer:2000}. Figure~\ref{fig:3tensor_unfolding} illustrates the unfolding of a third order tensor.

The unfolding makes it possible to rewrite the $n$-mode product of a tensor equivalently as a matrix-vector product. 
\begin{align}
	{}&\Acal = \Scal \times_1 \OP^{(1)} \ldots \times_N \OP^{(N)} \nonumber\\
	\Leftrightarrow\ &\Ab_{(N)} = \OP^{(N)} \cdot \Sb_{(N)} \cdot \left(\OP^{(1)}\otimes \ldots \otimes \OP^{(N-1)}\right)^\top \nonumber\\
	\Leftrightarrow\ &\VEC(\Ab_{(N)})=\left(\OP^{(1)} \otimes \ldots \otimes \OP^{(N)}\right) \cdot \VEC(\Sb_{(N)}) \label{eq:vectorized_nmode}
\end{align}
Here, $\VEC$ denotes the operator that stacks the columns of a matrix below one another and $\otimes$ denotes the Kronecker product of matrices. The first reshaping is obtained by unfolding $\Acal$ and $\Scal$ along the $N^\mathrm{th}$ mode, while the second follows from the definition of the Kronecker product.

We refer to a large, structured cosparse analysis operator $\OP$ that can be represented as a concatenation of smaller operators $\{\OP^{(1)},\ldots,\OP^{(N)}\}$ as a \emph{separable} operator.

\subsection{Derivation of the Cost Function}
With the introduced tensor operations we are able to define a way to apply a cosparse analysis operator to a multidimensional signal $\Scal$ recorded as an $N$-tensor. 
Let $\{\Scal_i\}$, $i=1,\ldots,T$ be a set of training signals where each signal is a tensor of order $N$.
The optimal separable cosparse analysis operator for this set is learned by solving the learning problem
\begin{equation}\begin{split}
\label{eq:def_sparsepen}
	\arg\!\!\! \min_{\OP^{(i)} \in \mathfrak{C}_i} \tfrac{1}{T} \sum_{j=1}^T g\left(\Scal_j \times_1 \OP^{(1)} \ldots \times_N \OP^{(N)}\right)^2
\end{split}\end{equation}
with the sparsity promoting function
\begin{equation}
\label{eq:sparsity_measure}
	g(\Acal) \coloneqq \sum_{k}\log\left( 1 + \nu \alpha_k^2 \right).
\end{equation}
The parameter $\nu>0$ serves as a weighting factor to control the sparsity, and the sum is taken over all entries $\alpha_{k}$ of the tensor $\Acal$.
There are two things to point out in the design of Equation~\eqref{eq:def_sparsepen}. 
First, in the proposed optimization problem we minimize the square of $g(\cdot)$. 
As shown in \cite{hawe:tip13} this leads to a balanced minimization of expectation and variance of the samples' sparsity.
Second, the minimization is performed over a constraint set $\mathfrak{C}_i$ which prevents the trivial solution of the given problem. The role of this constraint set will be discussed in the following.

As shown in \eqref{eq:vectorized_nmode}, the $n$-mode product which is used in the cost function can be rewritten as a matrix-vector-product. The Kronecker product of the small operators that appears in \eqref{eq:vectorized_nmode} can be interpreted as a large cosparse analysis operator with additional structure which is applied to the vectorized signal.
The algorithm we propose is based on the work by the authors in \cite{hawe:tip13} and extended to a separable case.
In \cite{hawe:tip13}, the authors enforce certain properties on their unstructured operator $\widehat \OP$ via regularizers. 
They demand that
\begin{enumerate}[label=(\roman*), leftmargin=*]
	\item The rows of $\widehat \OP$ have unit Euclidean norm.
	\item The operator $\widehat \OP$ has full rank, i.e.\ it has the maximal number of linear independent rows.
	\item The rows of the operator $\widehat \OP$ are not trivially linearly dependent, i.e.\ $\omegab_{k,:} \neq \pm \omegab_{l,:}$ for $k \neq l$.
\end{enumerate}
As stated in \cite{hawe:tip13} condition (i) is realized by the oblique manifold
%
\begin{equation*}
	\widehat{\OB}(\coeffdim,n) \!=\! \big\{ \OP \in \R{\coeffdim \times n} \,:\, (\OP \OP^\top)_{ii} = 1,\, i=1,\ldots, \coeffdim \big\}.
\end{equation*}
%
%
In the following, we show that for a structured cosparse analysis operator enforcing these constraints on the operator components is equivalent to enforcing them on the large operator. It should be pointed out that the cosparse analysis operators for the separate modes are actually the transposed of an element of the oblique manifold of appropriate size. 

It follows from the definition of the Kronecker product that if the rows of $\OP^{(1)}$ and $\OP^{(2)}$ have unit norm, then the rows of $\OP^{(1)} \otimes \OP^{(2)}$ have unit norm as well.
Furthermore, the rank of the matrix resulting from Kronecker product $\OP^{(1)} \otimes \OP^{(2)}$ is the product of the rank of $\OP^{(1)}$ and $\OP^{(2)}$. 
Finally, if the operators $\OP^{(1)}, \OP^{(2)}$ do not exhibit trivially linearly dependent rows, then neither does $\OP^{(1)} \otimes \OP^{(2)}$.
By induction the above mentioned properties can be extended to the Kronecker product of finitely many operators.
Hence, the properties demanded in \cite{hawe:tip13} are fulfilled for $\OP^{(1)} \otimes \ldots \otimes \OP^{(N)}$ if the conditions (i), (ii), and (iii) hold for each operator component $\OP^{(i)}, i = 1,\ldots,N$.
To realize the full rank and the incoherence constraints, we adopt the regularizers
\begin{align*}
	h\big(\OP^{(i)}\big) &= - \tfrac{1}{n_i \log(n_i)}\log \det \left(\tfrac{1}{\coeffdim_i}(\OP^{(i)})^\top \OP^{(i)}\right),\\
	r\big(\OP^{(i)}\big) &= - \sum_{k<l} \log \left( 1 - \left((\omegab_{k,:}^{(i)})^\top (\omegab_{l,:}^{(i)})\right)^2 \right).
\end{align*}

The sum of these log-barrier functions is itself a logarithmic barrier functions which ensures that $\OP^{(i)}$ lies within the intersection of the two proposed constraint sets.
With the corresponding weights $\kappa$ and $\mu$ which control the importance of the corresponding constraint, we are able to formulate the optimization problem
\begin{align}
	\arg\min_{\OP^{(i)}} \tfrac{1}{T} \sum_{j=1}^T &g\left(\Scal_j \times_1 \OP^{(1)} \ldots \times_N \OP^{(N)}\right)^2\nonumber\\[-0.75em]
	{}&+ \kappa \sum_{j=1}^N h\big(\OP^{(j)}\big) + \mu \sum_{j=1}^N r\big(\OP^{(j)}\big)\nonumber\\
	\text{subject to: }\quad &\OP^{(i)} \in \widehat{\OB}(\coeffdim_i,n_i),\quad i=1,\ldots,N. \label{eq:AOL_problem}
\end{align}
%

This optimization task is tackled using a geometric conjugate gradient method based on the one introduced in \cite{hawe:tip13}. 
Our algorithm is an extension in the sense that it is designed to work on the product of oblique manifolds. 
The line search required in the CG algorithm is performed using the non-monotone line search strategy proposed in \cite{ls:zhang:2004}.

\section{Multidimensional Signal Reconstruction}
\label{sec:invprob}

Multidimensional signals can be found in many applications ranging from video, over hyper-spectral imaging to 3D models, medical images, and others. 
However, even within one signal class each dimension can represent a completely different physical meaning. 
While in hyper-spectral data the third dimension encodes the spectral component of a scene, volumetric data typically encodes spatial information.
In this work we focus on three-dimensional data with a homogeneous physical meaning in all dimensions. An example for data of this kind is volumetric MRI data on which we evaluate our proposed scheme by conducting several signal reconstruction experiments.

For this purpose, we first extract a training set of {$T = 20\,000$} non-flat and normalized tensors of size $\Scal_i \in \R{5 \times 5 \times 5}$ from a synthetic, unbiased MRI brain atlas \cite{Fonov2009,Fonov2011}, which is available online\footnote{http://www.bic.mni.mcgill.ca/ServicesAtlases/ICBM152NLin2009}. The parameter $\nu$ in \eqref{eq:sparsity_measure} is set to $\nu = 1\,000$, while the weightings of the constraints read $\kappa = 500$ and $\mu = 0.5$. After learning, we use the operators each of size $\OP^{(i)} \in \R{6 \times 5},\, i = 1,2,3$ to regularize the reconstruction formulated as an inverse problem.
However, since the operators only act on local signal cubes, we first introduce the operator ${\bf{\Pi}}_{(r,c,k)}(\cdot)$ that extracts a small tensor of appropriate size centered at location $(r,c,k)$ from the entire volume. Given the measurements $\yb = {\bf{\Phi}}(\Vcal) + \eta \in \R{J_1 \cdot J_2 \cdot J_3}$ arranged as a vector, where ${\bf{\Phi}}(\Vcal)$ models the measurement process and $\eta$ constitutes some noise, we finally state the reconstruction of the original volumetric signal $\Vcal \in \R{I_1 \times I_2 \times I_3}$ as the inverse problem
\begin{equation}\begin{split}
\label{eq:sig_rec}
	\Vcal^\star & \in  \arg\min_{\Vcal}  \,\, \tfrac{1}{2} \| {\bf{\Phi}}(\Vcal) - \yb \|_{2}^{2} \\
	&+ \tfrac{\lambda}{M} \!\!\! \sum_{(r,c,k)} \!\!\! g\big( {\bf{\Pi}}_{(r,c,k)} ( \Vcal )  \times_1 \OP^{(1)} \times_2 \OP^{(2)} \times_3 \OP^{(3)}\big).
\end{split}\end{equation}
In all our experiments, we employ the same sparsity promoting function $g(\cdot)$ as given in \eqref{eq:sparsity_measure}. The parameter $\lambda \in \R{+}$ balances between the data fidelity and the regularization and $M$ serves as a normalization factor. Optimization is performed via a conjugate gradient approach. Our algorithm is named MAOL (Multidimensional Analysis Operator Learning).
\begin{table}[t]
\small
\renewcommand{\arraystretch}{1.3}
\caption{MRI volume data reconstruction from measurements corrupted by additive white Gaussian noise with standard deviation $\sigma_{\text{noise}}$. Achieved PSNR in decibels, $t$ in minutes.} 
\label{tab:denoising}
\centering
\begin{tabular}{c|c||c|c||c|c}
\hline
Method    &  $\sigma_{\text{noise}}$ & \textit{PSNR} & \textit{MSSIM} & $t_{\text{learn}}$ &  $t_{\text{rec}}$   \\
\hline
AKSVD     &       5                  &   38.41       &   0.968       &   980       &   840    \\
          &       15                 &   28.83       &   0.798       &             &          \\ \hline
MAOL      &       5                  &   38.55       &   0.971       &   18        &   12      \\
          &       15                 &   30.64       &   0.851       &             &           \\
\hline
\end{tabular}
\vspace{-9pt}
\end{table}

In our first experiment we evaluate the influence of preserving the multidimensional structure compared to the standard approach of vectorizing each tensor. For this purpose we conducted a simple denoising experiment.
The second experiment illustrates the flexibility of our method which is applicable to a wide range of image processing tasks. A prominent linear inverse problem in medical imaging is the reconstruction of MRI images from partial Fourier measurements. Reducing the amount of measurements and thus reducing the scan time for MRI is highly desired for clinical throughput and convenience for patients.
Based on the theory of Compressive Sensing (CS), many algorithms have been proposed in the literature so far that exploit the sparsity of this type of data. Typically, sparsity is measured in a fixed analytic transform domain like in the wavelet basis, while additional priors like total-variation further enhance the reconstruction quality. 
However, recently learning approaches have made their way into this field due to the fact that they adapt the structure of the signal more accurately.
The original MRI data has been scaled such that its intensity values lie in the range $[0,255]$. The performance is measured in terms of PSNR and the mean structural similarity (MSSIM) averaged over all slices in the whole volume. All experiments are implemented in \mbox{MATLAB} and executed on a standard desktop PC.

\section{Numerical Experiments}
\label{sec:exp}
Denoising aims at reconstructing a clean signal version from noise corrupted data, i.e.\ we have ${\bf{\Phi}}(\Vcal) = \VEC(\Vcal)$ and $\eta$ is additive white Gaussian noise (AWGN) with standard deviation $\sigma_{\text{noise}}$. The volumetric MRI test data is a representative region of size $70 \times 70 \times 70$ voxels from the T1 weighted BrainWeb phantom available online\footnote{http://brainweb.bic.mni.mcgill.ca/brainweb/} \cite{Cocosco97,Kwan1999} which has been artificially corrupted by AWGN. We chose the Analysis \mbox{K-SVD} (AKSVD) algorithm proposed in \cite{rubinstein2013analysis} for comparison and learned a single operator $\OP_{\mathrm{KSVD}} \in \R{216 \times 125}$, i.e.\ the analyzed vector has the same dimension as ours. The operator $\OP_{\mathrm{KSVD}}$ is directly learned from $20\,000$ noise contaminated tensors from the data that has to be reconstructed. Table \ref{tab:denoising} summarizes the results for two different choices of $\sigma_{\text{noise}}$ as well as the computation time for the operator learning $t_{\text{learn}}$ and the reconstruction $t_{\text{rec}}$. We assume $\sigma_{\text{noise}}$ to be known and set $\lambda = 200\sigma_{\text{noise}}$.
%
\begin{table}[t]
\small
\renewcommand{\arraystretch}{1.3}
\caption{Compressive Sensing MRI volume reconstruction from noiseless radial sampled measurements with an undersampling rate of 20\%. Achieved PSNR in decibels.}
\label{tab:csmri}
\centering
\begin{tabular}{c||c|c}
\hline
Method           & \textit{PSNR} & \textit{MSSIM}  \\
\hline
RecPF            &   26.06       &   0.788          \\ \hline
DLMRI            &   27.35       &   0.818          \\ \hline
MAOL             &   27.90       &   0.847          \\ 
\hline
\end{tabular}
\vspace{-3pt}
\end{table}
%

While the performance of the two algorithms in the range of moderate AWGN is comparable in terms of PSNR and MSSIM, the separable structure of our operator enables a significant saving of computation time compared to the vectorizing approach. 

\begin{figure}[t]
\centering
	\begin{subfigure}[b]{0.4\linewidth}
	\centering
	\includegraphics[width=0.8\linewidth]{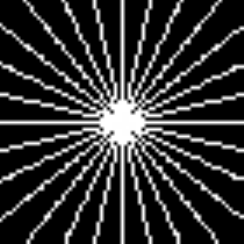}\subcaption{Sampling mask}\label{fig:subfig_sampling}
	\end{subfigure}
	\begin{subfigure}[b]{0.4\linewidth}
	\centering
	\includegraphics[width=0.8\linewidth]{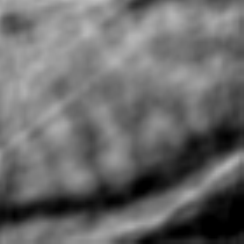}\subcaption{Zero-filling}
	\end{subfigure}\\ \vspace{3pt}
	\begin{subfigure}[b]{0.4\linewidth}
	\centering
	\includegraphics[width=0.8\linewidth]{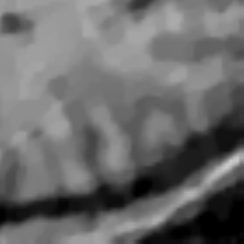}\subcaption{RecPF}
	\end{subfigure}
	\begin{subfigure}[b]{0.4\linewidth}
	\centering
	\includegraphics[width=0.8\linewidth]{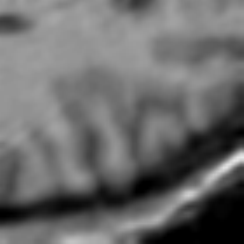}\subcaption{MAOL}
	\end{subfigure}
	\caption{Sampling mask and reconstruction results for (b) inverse FFT after zero-filling missing Fourier coefficients, (c) the RecPF algorithm, and (d) our proposed separable approach MAOL.\label{fig:res}}
	\vspace{-7pt}
\end{figure}

Regarding our second experiment, the underlying imaging process in MRI consists in measuring Fourier coefficients of a typically transversal slice rather than directly acquiring pixel values. Thus, the measurement process is modeled via ${\bf{\Phi}}(\Vcal) = {\bf{\Psi}}_{u}(\Vcal)$, where ${\bf{\Psi}}_{u}(\Vcal)$ performs a multi-slice 2D undersampled Fourier transform resulting in a vector containing the stacked measurements of each slice. The right choice of the chosen coefficients has a direct impact on the reconstruction quality. Here, we use radial sampling as illustrated in Fig.~\ref{fig:subfig_sampling} with an undersampling rate of 20\%. The weighting of the regularizer reads $\lambda = 1\,500$.
We compare our results to a standard approach (RecPF) \cite{Yang2010} that minimizes an error term consisting of a least squares data fitting term combined with total variation and wavelet sparsity regularization. Furthermore, we give the results of the algorithm proposed in \cite{bresler2011} (DLMRI) which utilizes an adaptively trained synthesis dictionary on each slice in the reconstruction process.
Results on the same noiseless volume as mentioned above are summarized in Table \ref{tab:csmri}.
%
The RecPF algorithm has the fastest execution time, however its accuracy suffers from the fixed analytical transformations. Both learning approaches DLMRI and MAOL achieve significantly better results, while our approach has a runtime eight times faster than DLMRI. Fig.~\ref{fig:res} shows 2D cross sections of the reconstructed volumes.
%
%

\section{Conclusion}
\label{sec:conc}
We provide a separable cosparse analysis operator learning framework for multidimensional signals called MAOL.
Due to the separable nature, the computational complexity of both learning and reconstruction is reduced which makes it possible to train the cosparse analysis operator on larger signal patches compared to methods that generate unstructured operators.
MAOL is designed using techniques from multilinear algebra and geometric optimization.
 The viability of our approach is demonstrated at the hand of experiments performed on volumetric MRI data. 

\bibliographystyle{IEEEbib}
\bibliography{refs}

\end{document}